\def\year{2018}\relax
\documentclass[letterpaper]{article} %DO NOT CHANGE THIS
\usepackage{aaai18}  %Required
\usepackage{times}  %Required
\usepackage{helvet}  %Required
\usepackage{courier}  %Required
\usepackage{url}  %Required
\usepackage{graphicx}  %Required
\usepackage{booktabs} % For formal tables
\usepackage{epsfig}
\usepackage{graphicx}
\usepackage{amsmath}
\usepackage{amssymb}
\usepackage{gensymb}

\usepackage{changes}
\usepackage{array}
\usepackage{enumerate}
\usepackage{adjustbox}
\usepackage{caption}
\usepackage{subcaption}
\usepackage{tabulary}
\usepackage{tabularx, blkarray}
\usepackage{eqparbox} 

\usepackage{color}
\usepackage{amsmath}
\usepackage{amssymb}
\usepackage{multirow, varwidth}
\usepackage{stfloats}
\usepackage{verbatim}
\makeatletter
\newif\if@restonecol
\makeatother

\usepackage[ruled,vlined]{algorithm2e}
\usepackage{xr}
\DeclareRobustCommand\onedot{\futurelet\@let@token\@onedot}
\def\onedot{. }
\def\eg{\emph{e.g}\onedot} 
\def\ie{\emph{i.e}\onedot}

\def\etal{\emph{et al}\onedot}

\newcommand{\x}{\mathbf{x}}
\newcommand{\y}{\mathbf{y}}

\usepackage{color}

\newcommand*{\affaddr}[1]{#1} % No op here. Customize it for different styles.

\begin{document}
%\title{A Deep Ranking Model for Spatio-Temporal Highlight Detection from a 360\degree Video}
\title{A Deep Ranking Model for Spatio-Temporal Highlight Detection  \\from a 360\degree~Video}
\author{%
	Youngjae Yu, Sangho Lee, Joonil Na, Jaeyun Kang, Gunhee Kim\\
	\affaddr{Seoul National University}\\
	{\tt\small \{yj.yu, sangho.lee, joonil\}@vision.snu.ac.kr, \{kjy13411\}@gmail.com, gunhee@snu.ac.kr} \\
%    \url{http://vision.snu.ac.kr/}
}

\maketitle
\begin{abstract}
We address the problem of highlight detection from a 360\degree~video by summarizing it both spatially and temporally.
Given a long 360\degree~video, we spatially select pleasantly-looking normal field-of-view (NFOV) segments from unlimited field of views (FOV) of the 360\degree~video, and temporally summarize it into a concise and informative highlight as a selected subset of subshots. %	as done in normal video summarization.
% Since there has been no prior work to address both spatial and temporal summarization of 360\degree~videos, 
We propose a novel deep ranking model named as \textit{Composition View Score} (CVS) model, 
which produces a spherical score map of composition per video segment, 
and determines which view is suitable for highlight via a sliding window kernel at inference.
To evaluate the proposed framework, we perform experiments on the Pano2Vid benchmark dataset~\cite{su:2016:ACCV} and our newly collected 360\degree~video highlight dataset from YouTube and Vimeo. 
Through evaluation using both quantitative summarization metrics and user studies via Amazon Mechanical Turk, 
we demonstrate that our approach outperforms several state-of-the-art highlight detection methods. %~\cite{su:2016:ACCV,gygli:CVPR:2016,yao:2016:CVPR}.
We also show that our model is 16 times faster at inference than AutoCam~\cite{su:2016:ACCV}, which is one of the first summarization algorithms of 360\degree~videos. 
\end{abstract}

\section{Introduction}
%------------------------------------------------------------------------------
% Figure 1: Key idea
\begin{figure}[t]
	\centering
	\includegraphics[width=0.46\textwidth]{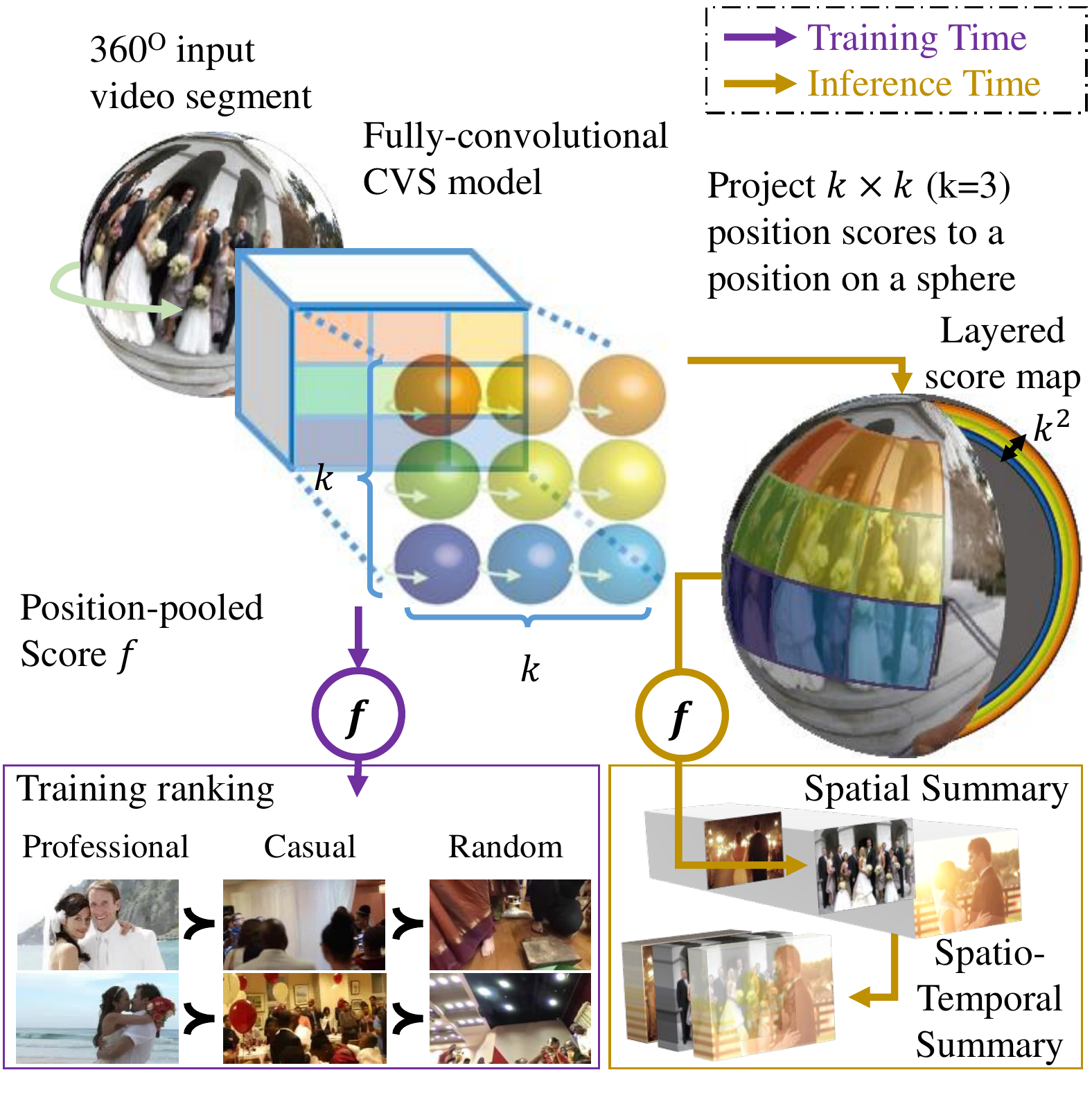}
	\caption{The intuition of the proposed \textit{Composition View Score} (CVS) model for highlight detection from a 360\degree~video. 
        For each 360\degree video segment,
        fully convolutional CVS model generates a layered spherical score map to determine which view is suitable for highlight. 
        It learns the fidelity of views from the professional videos of the same topic (\eg wedding) as reference, and at inference, a sliding window kernel computes the composition scores of views.
	}
	\label{fig:keyidea}
\end{figure}
%------------------------------------------------------------------------------

User-generated 360\degree~videos are flooding online, with emergence of virtual reality and active support by major social network platforms such as Facebook and YouTube. % content new 360\degree~cameras are flooding the market, including Samsung Gear 360, Ricoh Theta, 360Fly, to name a few. 
Since 360\degree~videos provide panoramic views of the entire scene, 
they free viewers not to get caught up in the intent of the videographer. % to maximize the viewer's experience.
However, without proper guidance, viewer experience can be severely handicapped, due to difficulty of understanding the entire content with limited human's field-of-view.
Therefore, like normal user videos, the highlight detection is also highly necessitated for much of online 360\degree~content, to quickly browse the overview of the content. 
One important difference of the 360\degree~video highlight detection is that the video summarization should be achieved both \textit{spatially} and \textit{temporally}. 
The spatial summarization selects pleasantly-looking normal field-of-view (NFOV) segments from unlimited field of views (FOV) of 360\degree~videos. Next, the temporal summarization generates a concise and informative highlight of a long video by selecting a subset of subshots.
%	as done in normal video summarization.
	
We address the problem of highlight detection from a long 360\degree~video by summarizing the video spatially and temporally. 
That is, we aim at selecting a pleasantly-looking NFOV within a 360\degree~FOV, and at the same time producing a concise highlight. 
To this end, we propose a novel deep ranking neural network, named as \textit{Composition View Score} (CVS) model. Given a 360\degree~video, the CVS model produces a spherical score map of composition. 
Here we use the term \textit{composition} to refer to a unified index to determine which view is suitable for highlight, considering all relevant properties such as the presence or absence of key objects, and the framing of objects. 
Based on the composition score map, we first perform \textit{spatial} summarization by finding out a best NFOV subshot per 360\degree~video segment, and then \textit{temporal} summarization by selecting $N$ top-ranked NFOV subshots as a highlight for the entire 360\degree~video.

Our approach has several noteworthy characteristics. % to the problem of 360\degree~highlight detection 
First, to learn the notion of good visual composition for highlight, we collect the online NFOV videos that are taken and edited by professional videographers or general users as positive reference.
Second, to quantify the difference between good and bad views, our deep ranking model learns the fidelity of views in the preference order of the professional NFOV video shots, normal users' NFOV shots, and randomly sampled NFOV shots from a 360\degree~video.
%\revise{\deleted{Third, to consider the relative positions of objects within a frame into the score computation, we leverage the position-pooling scheme like R-FCN~\cite{li:2016:NIPS}.}}
Third, to reduce the time complexity of the algorithm, we directly obtain a spherical score map for an entire 360\degree~video segment. It can substantially reduce redundant score computations of highly overlapping adjacent NFOV candidates, compared to other state-of-the-art spatial summarization methods of 360\degree~videos (\eg\cite{su:2016:ACCV,su:2017:CVPR}).

%As potential applications, our spatio-temporal ranking of highlights allows users to easily preview 360\degree~content, without continuously watching long 360\degree~videos and manually selecting their viewpoints out of hundreds of possible watching experiences. Moreover, our approach would relieve the burden of all time-consuming processes needed for 360\degree~video trailer production: from camera control to editing of the highlight, all of which can be automated.

For evaluation, we use the existing Pano2Vid benchmark dataset~\cite{su:2016:ACCV} for spatial summarization.
We also collect a novel dataset of 360\degree~videos from YouTube and Vimeo for spatio-temporal highlight detection. 
Our experiments show that our approach outperforms several state-of-the-art methods~\cite{su:2016:ACCV,gygli:CVPR:2016,yao:2016:CVPR} in terms of both quantitative summarization metrics (\eg mean cosine similarity, mean overlap, and mAP) and user studies via Amazon Mechanical Turk. % ~\cite{su:2016:ACCV} ~\cite{sun:2014:ECCV}

The major contributions of this work are as follows.

(1) To the best of our knowledge, our work is the first attempt to summarize 360\degree~videos both spatially and temporally for highlight detection.
To this end, we develop a novel  deep ranking model and collect a new dataset of 360\degree videos from YouTube and Vimeo. 

(2) We propose \textit{Composition View Score} (CVS) model, 
which produces a spherical composition  score map of composition per video segment of 360\degree videos, 
and determines which view is suitable for highlight via a sliding window kernel at inference.
Our framework is significantly faster at inference by reducing the redundant computation of adjacent NFOV candidates, which has been a serious problem in existing 360\degree video summarization models.

(3) For both Pano2Vid benchmark dataset~\cite{su:2016:ACCV} and our newly collected 360\degree highlight video dataset, 
our approach outperforms several state-of-the-art methods in terms of both quantitative metrics and user evaluation via Amazon Mechanical Turk. 
	
	% We propose a method for automated cinematography by transforming 360\degree~video into a single score map.
	% (2) We develop a variant of memory network model that can recover collective latent storyline among the videos for temporal summarization.
	% (3) With these two main works, we summarize the 360\degree~video in both space and time like a 2D video edited by a professional videographer. 

	%-------------------------------------------------------------------------
	
	% Relate work later, method and crawling first.
	\section{Related Work}
	\label{sec:related_work}

	%Recently, the former spatial summarization has been addressed in \cite{su:2016:ACCV,su:2017:CVPR}, named as the Pano2Vid task. 
	%However, our objective is to extend this summarization task into one more dimension of time domain. 

	\textbf{360\degree~video summarization}.
    There has been very few studies for summarization of 360\degree~user videos, except the \textit{AutoCam} framework proposed by Su \etal~\cite{su:2016:ACCV,su:2017:CVPR} and the deep 360 pilot proposed by Hu \etal~\cite{HuLin:2017:CVPR}. 
Compared to the AutoCam and deep 360 pilot method, our model has the following novelties in three respects.
First, in terms of problem definition, our work aims to summarize a 360-degree video both spatially and temporally, whereas the AutoCam and deep 360 pilot performs spatial summarization only. 
% We formulate it as a ranking problem that imposes the max-margin ranking loss in Eq.(3) (with video triplets of three different classes). it calculates a confidence score for extracting the spatial summary from every time step of a video.
Second, in the algorithmic aspect, the AutoCam uses a logistic regression classifier and the deep 360 pilot exploits recurrent neural networks to determine where to look at each segment; on the other hand, our model employs a deep fully-convolutional network. 
Third, our CVS model outperforms the AutoCam in terms of performance and computation time, which will be more elaborated in  our experiments.
For instance, our model greatly reduces the number of rectilinear projections and subsequent convolutional operations per score computation from 198 to 12 at the inference time.
 % for spatio-temporal summarization: for example, in terms of mAP for Wedding and MV datasets, CVS: (8.25, 8.86) vs. AutoCam: (6.51, 4.64). Moreover, our approach is much faster; 11 minutes compared to 178 minutes of the AutoCam as the average processing time of a one-minute 360-degree video.
% containing real world scenes. The framework identifies spatio-temporal glimpses that are worth capturing at every time step by using logistic regression.
% Since the AutoCam is one of few existing works, throughout this paper, we frequently compare our approach with AutoCam in methodology and evaluation.

	\textbf{Temporal video summarization}.
	Temporal video summarization provides a compressed abstraction of the original video while maintaining its key content~\cite{truong:2007:video}. There are many criteria for determining key content, such as visual attention~\cite{ejaz:2013:efficient,borji:2013:state},  importance or interestingness~\cite{gygli:2014:ECCV,gygli:2013:interestingness,fu:2014:interestingness} and diversity/non-redundancy~\cite{liu:2010:hierarchical,zhao:2014:CVPR}. Recently, many approaches use web-image priors for selecting informative content, assuming that images of the same topic often capture key events in high quality~\cite{kim:2014:CVPR:videostory,khosla:2013:CVPR,song:2015:CVPR}. As another direction, several methods use supervised-learning dataset including human-annotated summaries~\cite{gong:2014:diverse,gygli:2015:CVPR} to learn balanced score functions between interestingness and diversity. 
	
	\textbf{Video highlights}.
Rather than capturing a variety of events, video highlight models mainly focus on the importance or interestingness to summarize videos. 
To measure the interestingness, several methods use hand-crafted features from various low-level image features, such as aesthetic quality, camera following, and close-ups of faces~\cite{gygli:2014:ECCV,lee:2012:discovering}. Another group of approaches exploit category-specific information to better define highlights~\cite{sun:2014:ECCV,potapov:2014:ECCV}. However, the scalability of domain-specific models is limited by the difficulty of collecting raw footages and their corresponding annotated videos.
Some methods, meanwhile, attempt to deal with inherent noise in the web crawling dataset. Gygli \etal~\cite{gygli:CVPR:2016} train a model to learn the difference between a good and a bad frame. Sun \etal~\cite{sun:2014:ECCV} train a latent linear ranking model to generate a summary of a raw video using its corresponding edited video available online.
	
	%-----------------------------------------------------------------------------------------------------------
	\begin{figure*}
		\centering
		\includegraphics[width=0.96\textwidth]{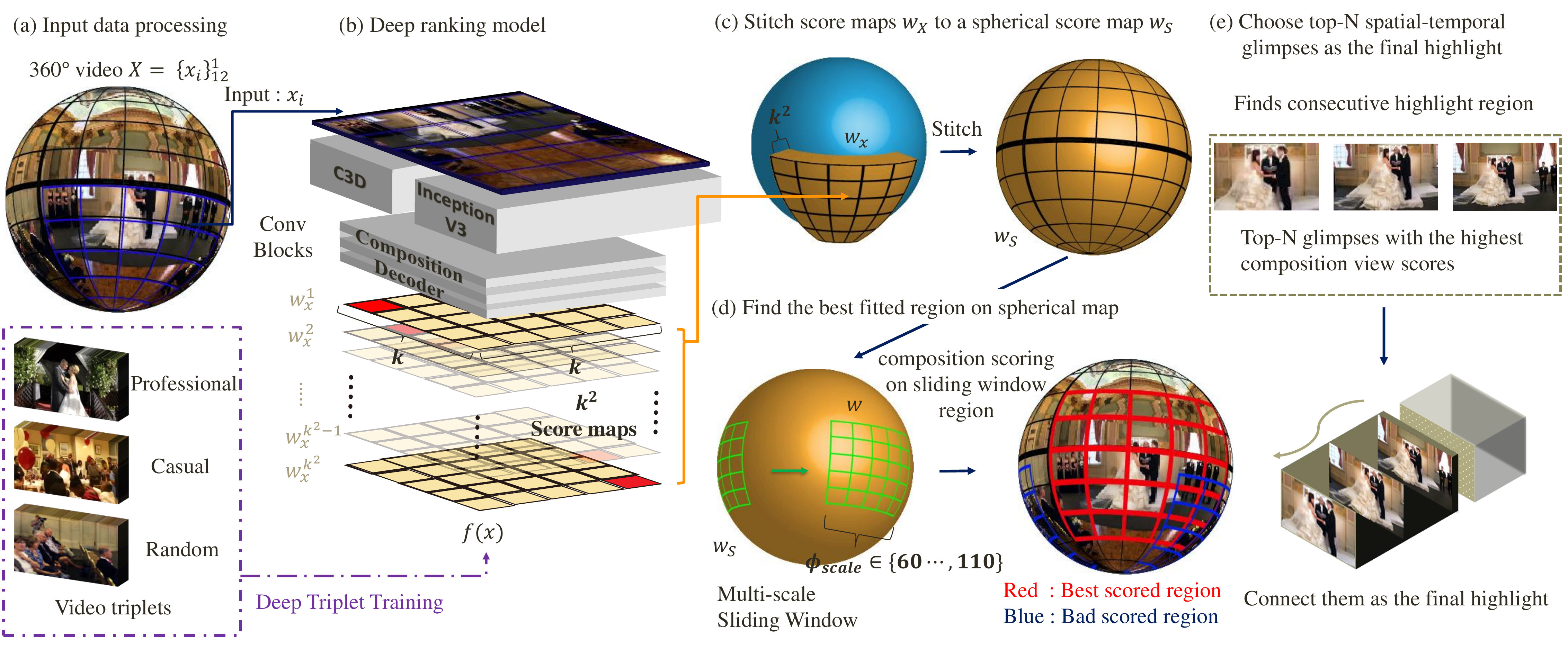}
		\caption{
			The overall framework for generating a spatio-temporal highlight using the Composition View Score (CVS) model.
		}
		\label{fig:archi}
	\end{figure*}
	%-----------------------------------------------------------------------------------------------------------

	\section{YouTube/Vimeo Dataset}
\label{sec:datacollection}

%%%% Table 1. Dataset statistics.
\newcolumntype{P}[1]{>{\centering\arraybackslash}p{#1}}
\begin{table}[t]
	\caption{Statistics of our new 360\degree~video highlight dataset.}
	\label{tbl:stats}
	\small
	\begin{center}
		\begin{tabular}{P{0.8cm}|c|P{1.45cm}|P{0.85cm}P{0.85cm} P{0.85cm}}
			\hline
			\multirow{2}{*}{Video} & \multirow{2}{*}{Topic} & \multirow{2}{*}{Type} & \multirow{2}{*}{\# video} & Total & mean \\
			& & & & (hour) & (minute) \\
			\hline
			360 \degree & wedding & \multirow{2}{*}{360\degree} & 62 & 54.8 & 53.1  \\ \cline{2-2} \cline{4-6}
			video & MV & & 53 & 17.2 & 19.5 \\
			\hline
			& \multirow{2}{*}{wedding} & professional & 755 & 87.1 & 6.9 \\ \cline{3-6}
			NFOV& & casual & 664 & 104.5 & 9.4 \\ \cline{2-6}
			video& \multirow{2}{*}{MV} & professional & 333 & 3.3 & 4.4  \\ \cline{3-6}
			& & casual & 654 & 47.5 & 0.6  \\ \cline{2-6}
			\hline
		\end{tabular}
		
	\end{center}
\end{table}
	
	We build a new dataset for the 360\degree~highlight detection task from YouTube and Vimeo. %consisting of 360\degree~user videos and professional/casual NFOV videos of the same topics.
	Its key statistics are outlined in Table~\ref{tbl:stats}.
	The \textit{professional} and \textit{casual} categories indicate NFOV videos by professional videographers and normal users, respectively.
	We select two popular topics: \textit{wedding} and \textit{music video}, which involve a large volume of both 360\degree~videos and NFOV videos that share the common storylines.
	Furthermore, the videos of these topics include multiple concurrent events, and thus are more interesting for spatio-temporal summarization.
	%We below describe the data collection process and the video format.

	\subsection{Video Format}
	\label{sec:preproc}

    We sample every video into 5 frames per second (fps). We then segment every video into a sequence of subshots (or segments) by using the content-based  subshot boundary detection of the \texttt{PySceneDetect}\footnote{https://github.com/Breakthrough/PySceneDetect.} library. % We extract samples from the resulting subshots with a random offset.
	
% to be compatible with the C3D feature extractor~\cite{tran:2015:ICCV}, which is pretrained on 16 consecutive frames of 5 fps videos from the UCF-101 dataset~\cite{soomro:2012:arxiv}. 
	
	% C3D architecture is a part of lower layer in CVS model described in subsection \ref{subsec:model_CVS}. 
	% For professional videos, we feed video frames directly to the model.
	% Unlike professional NFOV videos, 360\degree~videos require several preprocessing steps.
	In a 360\degree~video, there can be infinitely many view points at each time $t$, each of which can be determined from the latitude and longitude coordinates $(\theta,\phi)$ in the spherical coordinates. 
	Once we define a pair of $(\theta,\phi)$, we can generate an NFOV subshot from the 360\degree~video by a rectilinear projection with the viewpoint as a center.
	We set the size of each subshot to span a horizontal angle of 90\degree~with a 4:3 aspect ratio.
%    	As a result, we define each patch candidate by the camera principal axis direction $(\theta, \phi)$ and time $t$. %: \{ \Omega_{t,\theta,\phi} \equiv (\theta_t, \phi_t)\}\; in \;\Theta \times \Phi$.  
	The default display format for a 360\degree~video is usually obtained by equirectangular projection (\eg 2D world maps from the spherical Earth).
	%  which . The equirectangular projection is a simple map projection that displays a spherical 3d image in 2d. 
	This projection maps the meridians to vertical straight lines of constant spacing, and circles of latitude to horizontal straight lines of constant spacing. The resulting $\x$, $\y$ coordinates in the equirectangular format are:
%	\begin{align}
%	\label{eq:equirec_eq}
$	\mathbf x = (\phi-\phi_0)\cos\theta_1, \mathbf y = (\theta-\theta_1) $ % \nonumber \hspace{12pt} 
	% \phi = \dfrac{\x} {\cos\theta_1} + \phi_0 \hspace{12pt} \theta = \y + \theta_1
%	\end{align}
 where ${\theta}$ is the latitude, ${\theta_1}$ is the standard parallels, ${\phi}$ is the longitude, and ${\phi_0}$ is the central meridian of the map.
	
    Following \cite{su:2016:ACCV}, we use the term \textit{spatio-temporal glimpses} (or simply \textit{glimpses}) as a five-second NFOV video clip sampled from a 360\degree~video, from the camera principal axis direction $(\theta, \phi)$ at time $t$. %(rectilinearly projected)
    Therefore, the spatial summarization reduces to selecting a best ST-glimpses for a sequence of 360\degree~video segments.

\section{Approach}
\label{sec:approach}

Figure~\ref{fig:archi} illustrates the overall framework to generate a spatio-temporal highlight for a long 360\degree~video. 
The input of our framework is a long 360\degree~video in a form of a sequence of video segments $V = \{v_1, \cdots, v_T\}$, where each segment $v_t$ consists of spherical frames of 5 seconds. % at 16 time steps. 
The output is a sequence of highlight NFOV video subshots $S = \{s_1, \cdots, s_N\}$, where each video subshot $s_i$ consists of video frames of 5 seconds, 
and $N$ is a user parameter to set the length of the final highlight video. % and the output $S$ takes the form of video skimming.

\subsection{The Composition View Score Model}
\label{subsec:model_CVS}

\textbf{Model architecture}.
The \textit{Composition View Score} (CVS) model is the key component  of our framework. 
It computes a composition score for any NFOV spatio-temporal glimpse $\mathbf x$ sampled from a 360\degree~video. As shown in Figure~\ref{fig:archi}(b), the model is fully convolutional; it consists of feature extractors followed by a learnable deep convolutional decoder.

We represent a spatio-temporal glimpse $\mathbf x$ using two feature extractors, C3D~\cite{tran:2015:ICCV} pretrained on the UCF-101~\cite{soomro:2012:arxiv} for motion description, and Inception-v3~\cite{szegedy:2016:CVPR} pretrained on the ImageNet dataset~\cite{imagenet:2015:ijcv} for frame description.
For C3D features, we obtain the conv4b feature map $\mathbf x_m \in \mathbb R^{14\times14\times512}$ using a 112$\times$112 resized glimpse.
For Inception-v3 features, we use the mixed\_6c feature map $\mathbf x_f \in \mathbb R^{14\times14\times768}$ from a 260$\times$260 sized glimpse.
Finally, we stack $\mathbf x_m$ and $\mathbf x_f$ to be $\mathbf x_s \in \mathbb R^{14\times14\times1280}$, which is the input of our deep convolutional decoder.
	
The decoder consists of five convolutional layers, as shown in Table~\ref{tbl:decoder_archi}. % \revise{and is inspired by the position-pooling operation of R-FCN~\cite{li:2016:NIPS}.}.
It produces a set of $k^2$ score maps $\mathbf w_x = \{\mathbf w_x^1,\mathbf w_x^2 \cdots, \mathbf w_x^{k^2}\} \in \mathbb R^{k\times k\times k^2}$,
whose $(k \times i + j)$ channel $\mathbf w_x^{k \times i + j}$ encodes the composition for the $(i,j)$ position in a $k \times k$ grid.
% We then compute the composition score of $\mathbf x$ on these $k^2$ score maps.
We divide $\mathbf x$ into $k \times k$ bins by a regular grid, \ie, $\mathbf x = \{ x_{1,1}, \cdots,  x_{i,j}, \cdots,  x_{k,k}\}$, and aggregate all of position-wise scores using a Gaussian position-pooling:
\begin{align}
\label{eq:composition_score}
&f(\mathbf x) = \sum_{i,j} \sum_{l,m} \kappa(l-i) \kappa(m-j)~\mathbf w_{x}^{c(k,l,m)}(i,j|\mathcal{M}), \\
&\mbox{ where } \kappa(u) = \frac{\mbox{exp}(-u^2/2h^2)}{\sqrt{2\pi} h}, c(k,l,m) = k \times l + m, \nonumber
%\label{eq:spatial_attention_softmax}
\end{align}
\noindent for $i, j, l, m \in \{0, 1, \cdots, k-1\}$.
Here, $\kappa$ is a Gaussian kernel, $h$ is the kernel bandwidth, and $\mathcal{M}$ denotes all the CVS model parameters. We set default $k$ to 5.

It is partly inspired by the position-pooling of R-FCN~\cite{li:2016:NIPS}.
However, our position-pooling softly encodes all score maps according to their relative position in the $k \times k$ grid using the Gaussian kernel, while R-FCN pools only over the $(i, j)$-th score map.
This enhances the scale invariance of our model, and produces a better score by considering the surrounding context within the regular grid.

%-------------------------------------------------------------------------------------------
% Table : The architecture of the composition decoder.
\newcolumntype{P}[1]{>{\centering\arraybackslash}p{#1}}
\newcolumntype{L}[1]{>{\arraybackslash}p{#1}}
\begin{table}[t]
	%\footnotesize
	\centering
	\caption{
		The architecture of the composition decoder. We set $k$ to 5, and use stride 1 and no zero-padding for all layers. 
	}
	\label{tbl:decoder_archi}
	\begin{tabular}{c|c|c}
		\hline
		layer name & output size  & kernel size /  \# of channels   \\ \hline    %~\cite{su:2016:ACCV}
		conv1   & 12$\times$12$\times$512 &  3 $\times$ 3 / 512    \\ 
		conv2   & 10$\times$10$\times$512 &  3 $\times$ 3 / 512    \\ 
		conv3   & 8$\times$8$\times$1024 &  3 $\times$ 3 / 1024    \\ 
		conv4   & 5$\times$5$\times$2048 &  4 $\times$ 4 / 2048    \\ \hline
		pos\_map   & 5$\times$5$\times$25 &  1 $\times$ 1 / 25    \\ \hline		
	\end{tabular}	
\end{table}
%-------------------------------------------------------------------------------------------

\textbf{Training with video triplets}.
We pose the 360\degree~highlight detection as a ranking problem, in which the CVS model selects the NFOV glimpse $\mathbf x$ with the highest composition view score $f(\mathbf x)$ of Eq.(\ref{eq:composition_score}), from many possible NFOV glimpses in a 360\degree~frame. It is different from the AutoCam \cite{su:2016:ACCV} formulating as a binary classification problem using logistic regression, 
which is limited to rank a large number of candidates finely. 
% We have found that the logistic regression has limit to rank candidates with a lot of capture-worthy candidates across temporal frames. 
% Since it is impossible to clearly set an absolute cut value to determine whether a view is suitable for highlight, 

\begin{figure}[t!]
	\centering
	\includegraphics[width=0.46\textwidth]{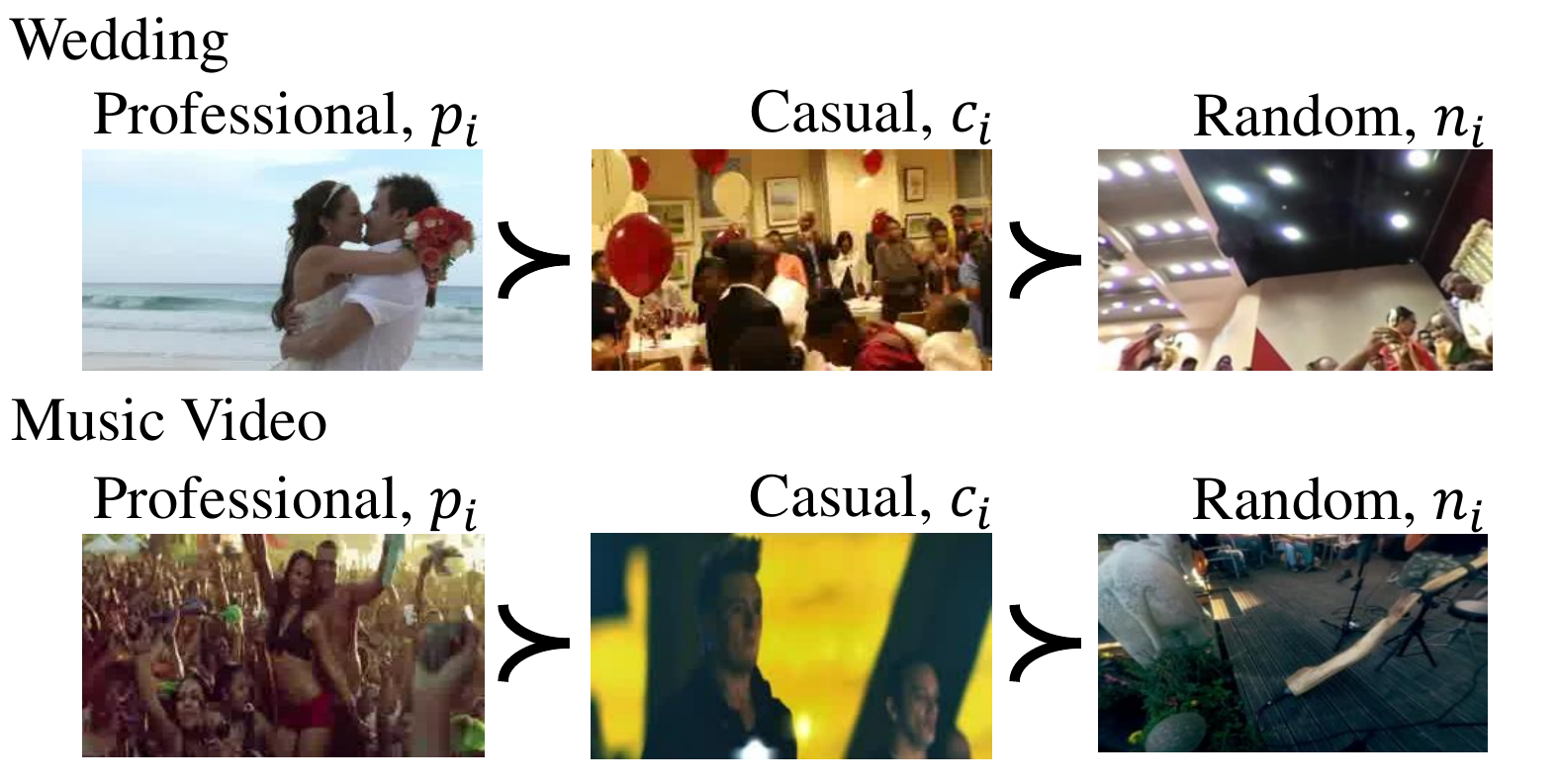}
	\caption{Examples of ranking order for professional, casual, and random NFOV glimpses about \textit{Wedding} and \textit{MV} topics.}
	\label{fig:rank_order}
\end{figure}

Our goal is to learn the CVS architecture that assigns a higher score to a view with good composition suitable for highlight. 
Rather than defining the goodness of composition with hand-crafted heuristics based on cinematography rules~\cite{arev:2014:automatic,heck:2000:towards,gleicher:2002:framework,heck:2007:virtual}, 
we leverage a data-driven approach; we collect professional NFOV videos and normal users' casual NFOV videos for the same topic, and use them as positive references of view selection. 
Since there are not many professional videos, we also exploit normal users' videos, although they are not as good as professional ones. 
During data collection, we observe that the quality gaps between professional videos and normal users' casual videos are significant. 
Thus, as highlight exemplars, we rank professional NFOV videos higher than the casual ones to correctly quantify the quality differences among the positive samples.
Assuming that a randomly selected view is likely to be framed badly, 
we regard a randomly cropped glimpse from a 360\degree~video as negative samples.  % classify spatio-temporal glimpses
As a result, we define the ranking constraints over the training dataset $\mathcal{D}$, which consists of video triplets of three different classes as shown in Figure \ref{fig:rank_order} as follows:
\begin{align}
\label{eq:ranking_eq}
f(p_i) \ \succ \ f(c_i)  \ \succ \ f(n_i), \hspace{6pt} \forall ~(p_i,c_i,n_i) \in \mathcal{D},
\end{align}
\noindent where $p_i$, $c_i$, $n_i$ indicate a professional, casual, and negative sample (\ie a video segment), respectively. 

To impose the ranking constraints, we train the CVS model using the following loss.
In particular, we assign different weights to different type pairs of video classes:

\begin{align}
\label{eq:max-margin_loss}
\mathcal{L}_i &= \alpha \max(0, f(c_i) - f(p_i) + 1)  \\ \nonumber
&+(1-\alpha) \max(0, f(n_i) - f(c_i) + 1), \\ \nonumber
\end{align}

\noindent where $\alpha \in [0, 1]$ is a hyperparameter; we set $\alpha=0.3$ in our implementation.
The final objective is the total loss over the dataset $\mathcal{D}$ combined with a $l_2$ regularization term:
\begin{align}
\label{eq:max-margin_objective}
\mathcal{L} = \sum_i \mathcal{L}_i+\lambda||\mathcal{M}||_{F}^{2}.
\end{align}
\noindent where $\lambda$ is a regularization hyperparameter.

%------------------------------------------------------------------------------
% Figure ?: Training pair sample for highlight
%------------------------------------------------------------------------------

\subsection{Inference}
\label{subsec:inference}

The goal of inference is to select a best NFOV glimpse for a given 360\degree~video segment. 
By repeating this process and connecting the selected glimpses, we can construct a highlight summary.
For efficient inference, we first compute a 360\degree composition score map for each 360\degree video segment using the learned CVS model. 
We then perform sliding window search over the  360\degree score map to select a highlight view. 

\textbf{360\degree composition score maps}.
Note that there has been proposed no CNN architecture that takes a 360\degree~video segment as input and produces a 360\degree~score map as output.
Therefore, we first propose an approximate procedure to obtain a spherical score map for a 360\degree video segment $v_t$ at time $t$ as follows.
We divide $v_t$ into 12 spatio-temporal glimpses $\mathbf X_t$ by discretizing viewpoints at longitudes $\mathbf{\phi} \in \Phi = \{0\degree,90\degree, 180\degree, 270\degree \} $ and at latitudes $\mathbf{\theta} \in \Theta = \{0\degree, \pm 67.5\degree \}$. 
Since each glimpse spans 90\degree~in the latitude axis with a  4:3 aspect ratio, $\mathbf X_t$ does not nearly overlap with one another while covering the entire spherical view.

Next, we transform every spatio-temporal glimpse $\mathbf x \in \mathbf X_t$ to an NFOV using rectilinear projection, and feed it into the learned CVS model to compute its score map. 
However, one issue here is that since the CVS is convolutional, the information loss occurs near the boundaries of adjacent glimpses.
Likewise padding in the CNN, we adopt the following simple trick; when transforming every glimpse $\mathbf x$ to an NFOV using rectilinear projection, we enlarge it by $1/k$ over the original size, in order to create a score map $\mathbf w_{x'} \in \mathbb{R}^{(k+2) \times (k+2) \times k^2}$.
Later when we stitch  these 12 score maps $\mathbf w_{X, t}$ to make a single  360\degree~sphere score map $\mathbf w_{S, t}$, we cut out the padded boundary scores and use only $\mathbf w_x \in \mathbb{R}^{k \times k \times k^2}$ in the center. In this way, all the position scores are seamlessly wrapped in a sphere score map and without information loss around glimpse boundaries.

\textbf{Sliding window search}.
After generating the spherical score map $\mathbf w_{S, t}$, we find the best fitted highlight position using a flexible sliding window kernel.
As the sliding window scans on the score map, the composition score  at the window location is calculated by cropping the score map $\mathbf w$ of the area from $\mathbf w_{S, t}$.
With a fixed aspect ratio of 4:3, the sliding window can change its scale by varying its horizontal size, whose can be any $\mathbf{\theta_{scale}} \in [ 60\degree , 110\degree ]$.
In our experiments, we test three scales of sliding windows with the horizontal sizes of $(65.5\degree, 90\degree, 110\degree)$, and find the maximally scored position and scale.
%Note that by changing the horizontal angles the resultant scales NFOV candidates vary as well.

\textbf{Execution time comparison.}
% The key difference between AutoCam~\cite{su:2016:ACCV} and our framework lies in efficiency at inference time.
One critical issue of existing models for spatial summarization is time complexity at the inference stage. % (\ie selective a best NFOV view for a given 360\degree~video segment). 
For example, AutoCam~\cite{su:2016:ACCV} computes capture-worthiness scores for a fixed number of overlapping 198 glimpses per 360\degree~video segment, and chooses the best glimpse with the highest score.
In this process, AutoCam performs severely overlapped convolutional operations and too many rectilinear projections. 
This exhaustive approach requires high computation time; for example, 3 hours per 1 minute 360\degree~video as reported in~\cite{su:2016:ACCV}. 
To overcome this issue, our approach is to first precompute a non-overlapping 360\degree~score map, and then use a flexible sliding window to find a highlight NFOV at inference.
This approach greatly reduces the number of iterative rectilinear projections and CNN computations from 198 to 12, which subsequently curtails the processing time from 178 min to 11 min.
The comparison on actual execution time will be reported in Table~\ref{tbl:computation_cost} in our experiments. 

%Due to efficient pre-computed score map, our inference speed is much faster than the Pano2Vid\cite{su:2016:ACCV} that computes score on all overlapped spatio-temporal glimpse (198 patches). 

\subsection{Spatio-Temporal Summarization}

Since our framework is a ranking model, it is straightforward to extend our CVS model to spatio-temporal summarization. 
That is, rather than selecting a glimpse at every time step, we can rank all the glimpses over the entire video.
We first build a candidate set of glimpses, each of which is selected from every video segment of 5 seconds by using the composition view score and the smooth-motion constraint~\cite{su:2016:ACCV}, which enforces that the latitude and longitude difference between consecutive glimpses must be less or equal than 30\degree: $|\theta_t - \theta_{t-1}|, |\phi_t - \phi_{t-1}| \leq 30\degree$.
% We average the composition scores of selected glimpses at every subset.
We then choose top-$N$ glimpses with the highest composition scores, and connect them as a final highlight. $N$ is a user parameter for the highlight length.
Since the professional videos that our CVS model uses as training reference are often highlights edited by professional videographers, 
high-ranked glimpses become excellent highlight candidates.

\subsection{Implementation Details}
We initialize all training parameters using the Xavier initialization~\cite{glorot:2010:AISTATS} and insert the batch normalization~\cite{ioffe:2015:batch} prior to all convolutional layers. 
We optimize the objective in Eq.(\ref{eq:max-margin_objective}) using vanilla stochastic gradient descent with a mini-batch size of 16. 
Experimentally, we use leaky ReLU~\cite{maas:13:ICML} as non-linear activation, and 
% \deleted{, and apply the Adam optimizer~\cite{kingma:2015:ICLR} with setting $\beta_1 = 0.9$, $\beta_2 = 0.999$, and $\epsilon = 10^{-8}$}.}
set our initial learning rate as $0.001$ and divide it by 2 at every 8 epochs.

\section{Experiments}
\label{sec:experiments}
We evaluate the performance  of the proposed CVS model with two datasets.
First, using the Pano2Vid dataset~\cite{su:2016:ACCV}, we show that the CVS model improves the spatial summarization performance compared to several baseline methods. 
Second, using our novel 360\degree~video highlight dataset, we demonstrate that our framework achieves the state-of-the-art performance of generating spatio-temporal video highlights.

\subsection{Evaluation Metrics}

In the Pano2Vid dataset, human annotators provide multiple edited videos per 360\degree~video,
in which they label the center coordinates of the selected glimpses in the camera principal axes (\ie latitude and longitude) at each video segment.  
% The reason for making multiple human-edited videos per 360\degree~video is to allow for the possibility of various good trajectories.
Using the labeled coordinates as groundtruth, we compare the similarity between the human-made view trajectories and predicted trajectories. 
We use the metrics of \textit{mean cosine similarity} and \textit{mean overlap} as proposed in the Pano2Vid benchmark.

To quantify the highlight detection performance (\ie spatio-temporal summarization) in our dataset, we compute the \textit{mean Average Precision} (mAP)~\cite{sun:2014:ECCV}. As groundtruths, we add three different highlight annotations to each of 25 randomly sampled 360\degree test videos per topic. 
Four human annotators watch full 360\degree~videos and select top-$N$ salient glimpses as video highlight subshots, with at least a distance of 5 seconds between choices. % and less than 5\% of the total number of video segments. % by locating latitude and longitude coordinates like the Pano2Vid dataset.
% Each 360\degree~video is annotated by three different annotators. Our annotators were students aged between 23$\sim$27.

%-------------------------------------------------------------------------------------------
% Table Evaluation
\newcolumntype{P}[1]{>{\centering\arraybackslash}p{#1}}
\newcolumntype{R}[1]{>{\arraybackslash}p{#1}}
\begin{table}[t]
	%\footnotesize
	\caption{
		Experimental results of spatial summarization on the Pano2Vid~\cite{su:2016:ACCV} dataset. Higher values represent better performance in both metrics.
	}
	\small \centering
	\label{tbl:results_mc_ret}
	\begin{tabular}{L{3.5cm}|P{1.75cm}|P{1.75cm}}
		\hline
		
		\multirow{2}{*}{Methods   }        & Frame & Frame \\   %~\cite{su:2016:ACCV}
		& cosine sim & overlap            \\ \hline
		Center                                    & 0.572                   & 0.336              \\
		Eye-Level                                 & 0.575                   & 0.392              \\
		Saliency                                  & 0.387                   & 0.188              \\ \hline
		AutoCam (w/o stitching)                   & 0.541                   & 0.354              \\   %~\cite{su:2016:ACCV}
		AutoCam-stitch                          & 0.581                     & 0.389              \\    
		RankNet                                   & 0.562                   & 0.398              \\ %~\cite{gygli:CVPR:2016}
		TS-DCNN                                   & 0.578                   & 0.441              \\ \hline   %~\cite{yao:2016:CVPR}
		CVS-C3D                                   & 0.656                   & 0.554              \\
		CVS-Inception                             & 0.642                   & 0.545              \\
		CVS-Fusion                                & 0.701                   & 0.590              \\  \hline
		CVS-C3D-stitch                          & 0.774                   & 0.646              \\ 
		CVS-Inception-stitch                    & 0.768                   & 0.666              \\ 
		CVS-Fusion-stitch                       & \textbf{0.800}          & \textbf{0.677}     \\  \hline
		
	\end{tabular}

	\caption{Comparison of trajectory similarity after applying the stitch algorithm in \cite{su:2016:ACCV}.}
	\label{tbl:results_meteor}
	
	\begin{tabular}{ L{2.2cm}|P{1.2cm}|P{0.85cm}|P{1.2cm}|P{0.85cm}}
		\hline
		& \multicolumn{2}{c|}{Cosine sim} & \multicolumn{2}{c}{Overlap}   \\ \hline
		& Trajectory     & Frame          & Trajectory    & Frame          \\ \hline
		AutoCam-stitch   & 0.304          & 0.581          & 0.255         & 0.389          \\   %~\cite{su:2016:ACCV}
		CVS-C3D-stitch   & 0.524          & 0.774          & 0.503         & 0.646  \\
		CVS-Fusion       & \textbf{0.563} & \textbf{0.800} & \textbf{0.530}& \textbf{0.677}  \\
		\hline
	\end{tabular}

	\caption{Results of highlight detection on our 360\degree~video highlight dataset. Higher mAPs indicate better performance.}
	\label{tbl:results_mAP}
	\begin{tabular}{L{3.5cm}|P{1.75cm}|P{1.75cm}}
		\hline
		
		Methods & Wedding & MV \\ \hline
		Center & 7.88 & 5.90    \\ 
		% AutoCam & 6.51 & 4.64 \\ 
		RankNet & 11.98 & 11.65 \\
		TS-DCNN & 13.23 & 12.28 \\ \hline
		CVS-C3D & 16.32 & 12.15 \\
		CVS-Inception & 16.13 & 12.38 \\
		CVS-Fusion (pairwise) & 14.34 & 12.56 \\  \hline
		CVS-Fusion & \textbf{17.96} & \textbf{14.92} \\
		\hline
	\end{tabular}

\end{table}
%-------------------------------------------------------------------------------------------

%-------------------------------------------------------------------------------------------
\newcolumntype{P}[1]{>{\centering\arraybackslash}p{#1}}
\begin{table}[t]
	\centering
	%\footnotesize
	\small
	
	\caption{Comparison of computational costs between our CVS model and AutoCam~\cite{su:2016:ACCV}. The projected area is the total area of glimpses, expressed as multiples of the sphere area of a 360\degree frame. }
	
	\label{tbl:computation_cost}
	
	\begin{tabular}{L{1.4cm} |P{1.75cm}|P{1.75cm}|P{1.75cm}}
		\hline
		
		\small
		% Tip: To use auto align in VIM, press "vi[ga*&" somewhere within this block [[[
		&  Processing time & \# of ST-glimpses & Projected area      \\
		%\midrule
		\hline
		AutoCam                     & 178 min               & 198                  & $\times~4.5479$         \\  %\cite{su:2016:ACCV}
		\hline
		
		CVS                         & 11 min                & 12                   & $\times~1.96$              \\ 
		\hline
		
	\end{tabular}

\end{table}

%-----------------------------------------------------------------------------------------

\subsection{Baselines}

For performance comparison with our CVS model, 
we select (i) three simple baselines of spatial summarization used in \cite{su:2016:ACCV}, (ii) AutoCam~\cite{su:2016:ACCV} proposed for the original Pano2Vid task, and (iii) two state-of-the-art pairwise ranking model of deep neural networks for normal video summarization, RankNet~\cite{gygli:CVPR:2016} and TS-DCNN~\cite{yao:2016:CVPR}. 
All baselines share the same video feature representation by the C3D model~\cite{tran:2015:ICCV} pretrained on the UCF-101~\cite{soomro:2012:arxiv}, except the TS-DCNN as described below.

We do not consider the deep 360 pilot method~\cite{HuLin:2017:CVPR} as a baseline, because it is a supervised method that requires a center viewpoint per frame as a label for training.

\textbf{Center, Eye-level}~\cite{su:2016:ACCV}.
These two baselines are stochastic models tested in the Pano2Vid benchmark, that select spatio-temporal glimpses at each segment with the following preferences. 
The Center method samples the glimpses at $\theta = 0, \phi = 0$, and then performs a random motion with a Gaussian distribution. 
The Eye-Level method samples the glimpses at $\theta = 0$ with a fixed set of $\mathbf{\phi} \in \{0\degree, 20\degree, 40\degree, \cdots , 340\degree\} $.

\textbf{Saliency}~\cite{su:2016:ACCV}.
This baseline uses the graph-based saliency score~\cite{harel:2006:graph} to classify capture-worthy glimpses, 
and join the selected glimpses using the motion constraint that AutoCam uses.

\textbf{AutoCam}~\cite{su:2016:ACCV}.
With C3D video features~\cite{tran:2015:ICCV}, AutoCam uses the logistic regression to classify which glimpses are worth to capture. It generates a summary video in the following two steps: (i) sampling the best scored glimpses at each video segment, and (ii) stitching the glimpses based on both the capture-worthiness scores and a smooth motion constraint.
We denote the AutoCam using the both steps as AutoCam-stitch, while the one with the first step only as AutoCam. 

\textbf{RankNet}~\cite{gygli:CVPR:2016}.
RankNet is a deep pairwise ranking model used in Video2GIF task~\cite{gygli:CVPR:2016}.
Unlike our fully convolutional model, RankNet predicts a single score from the C3D features through additional consecutive fully connected layers.
We train the RankNet with the adaptive Huber loss as proposed in \cite{gygli:CVPR:2016}.

\textbf{Two-Stream DCNN}~\cite{yao:2016:CVPR}).
The TS-DCNN is a recent pairwise ranking model of spatio-temporal deep CNNs for ego-centric video highlight detection.
The spatial component using AlexNet \cite{krizhevsky:2012:NIPS} represents scenes and objects in the video by frame appearance, while the temporal counterpart using C3D~\cite{tran:2015:ICCV} conveys the motion dynamics.
Overall TS-DCNN is similar to the RankNet, although they use both spatial and temporal representation. Thus, it can be good comparison to our CVS-Fusion model.

%------------------------------------------------------------------------------
% Figure ?: projection comparison
\begin{figure}[t]
	\centering
	\includegraphics[width=0.46\textwidth]{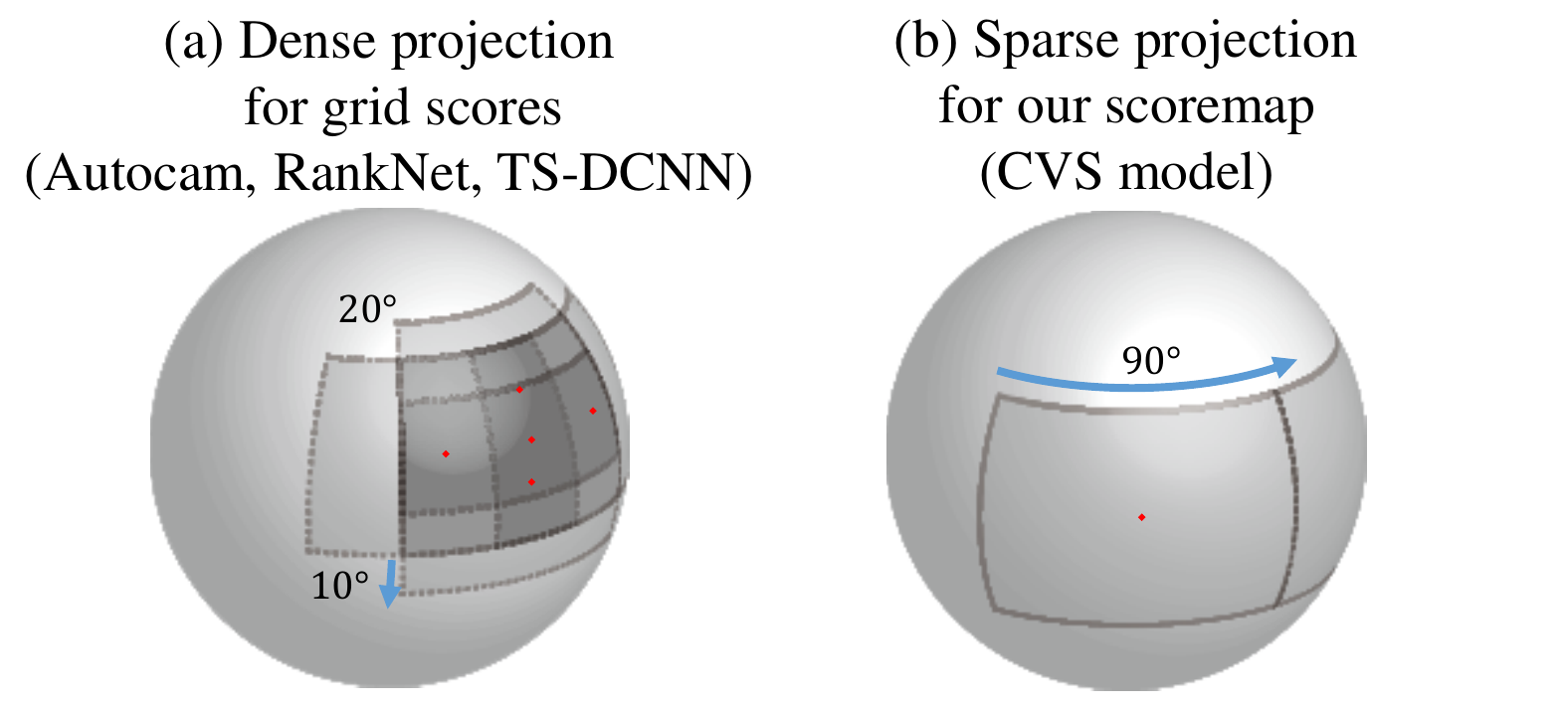}
	\caption{The visualization of (a) dense projection (198 glimpse) and (b) sparse projection (12 glimpse) by CVS score maps.  
	}
	\label{fig:projection_vis}
\end{figure}
%------------------------------------------------------------------------------

%------------------------------------------------------------------------------
% Figure ?: position score overlap
\begin{figure*}[t]
	\centering
\includegraphics[width=0.94\textwidth]{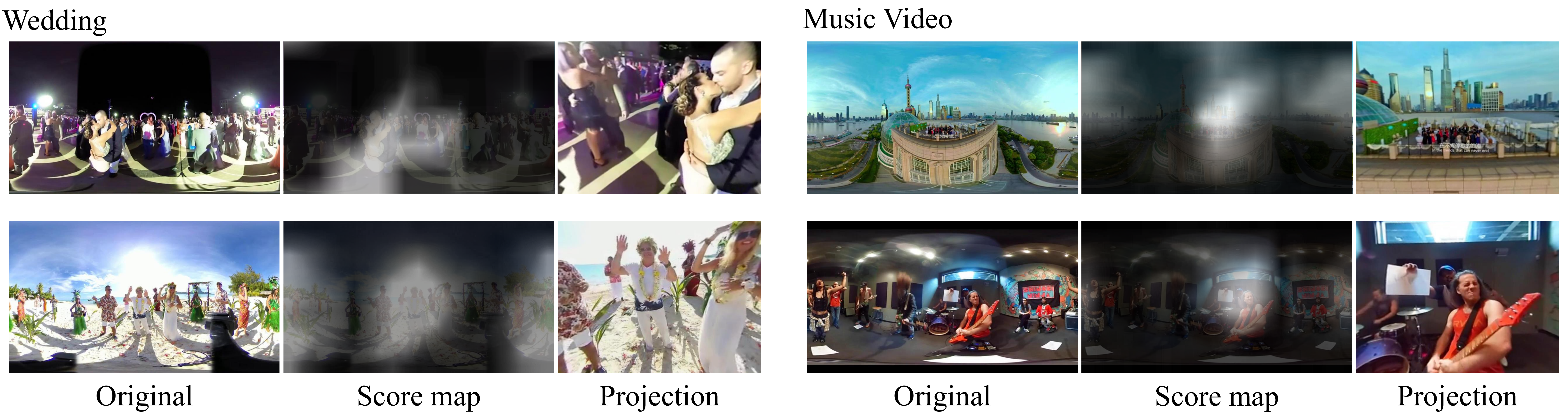} 

\caption{Examples of view selection. For a given spatio-temporal glimpse (left), we show the composition view score map (middle), and the projected NFOV with the highest score (right). 
The higher the view score is, the whiter it appears on the map.   
}
\label{fig:pos_overlap}
\end{figure*}
%------------------------------------------------------------------------------

%------------------------------------------------------------------------------
% Figure ?: Example score
\begin{figure*}[t]
	\centering
	\includegraphics[width=0.94\textwidth]{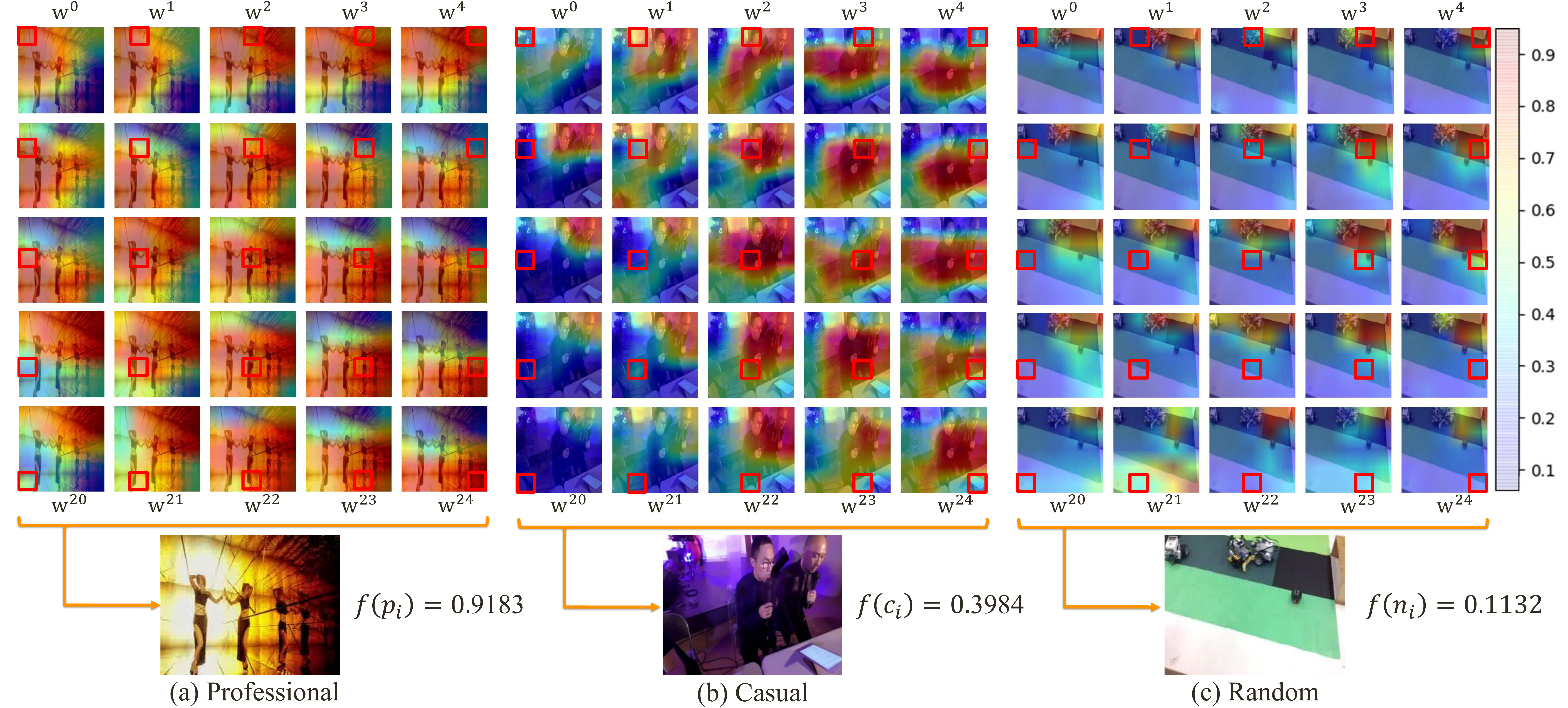} 
	\caption{Examples of 25 position score maps $\mathbf w_x$ for (a) a professional,  (b) a casual, and (c) a random glimpse in our Music Video (MV) highlight dataset. 
Our model successfully assigns higher scores to better views for the highlight.
	}
	\label{fig:example_attention}
\end{figure*}
%------------------------------------------------------------------------------

%-------------------------------------------------------------------------------------------
\newcolumntype{P}[1]{>{\centering\arraybackslash}p{#1}}
\begin{table}[t]
	\centering
	%\footnotesize
	\small
	\begin{tabular}{L{2.88cm}|P{2.25cm}|P{2.25cm}}
		\hline
		
		CVS-Fusion vs & Wedding & MV \\ \hline
		Center        & \textbf{68.0} \% (117/150) & \textbf{57.3} \% (86/150)    \\ 
		RankNet       & \textbf{67.3} \% (101/150) & \textbf{65.3} \% (98/150)    \\
		TS-DCNN       & \textbf{64.0} \% (96/150)  & \textbf{58.0} \% (87/150)   \\ 
		\hline

	\end{tabular}	
		\caption{AMT results for 360\degree~highlight detection.
		We show the percentages of turkers' votes for our method (CVS-Fusion) over baselines.}
	\label{tbl:results_AMT}

\end{table}
%-----------------------------------------------------------------------------------------

  %------------------------------------------------------------------------------
% Figure ?: projection comparison
\begin{figure*}[t]
	\centering
	\includegraphics[width=0.92\textwidth]{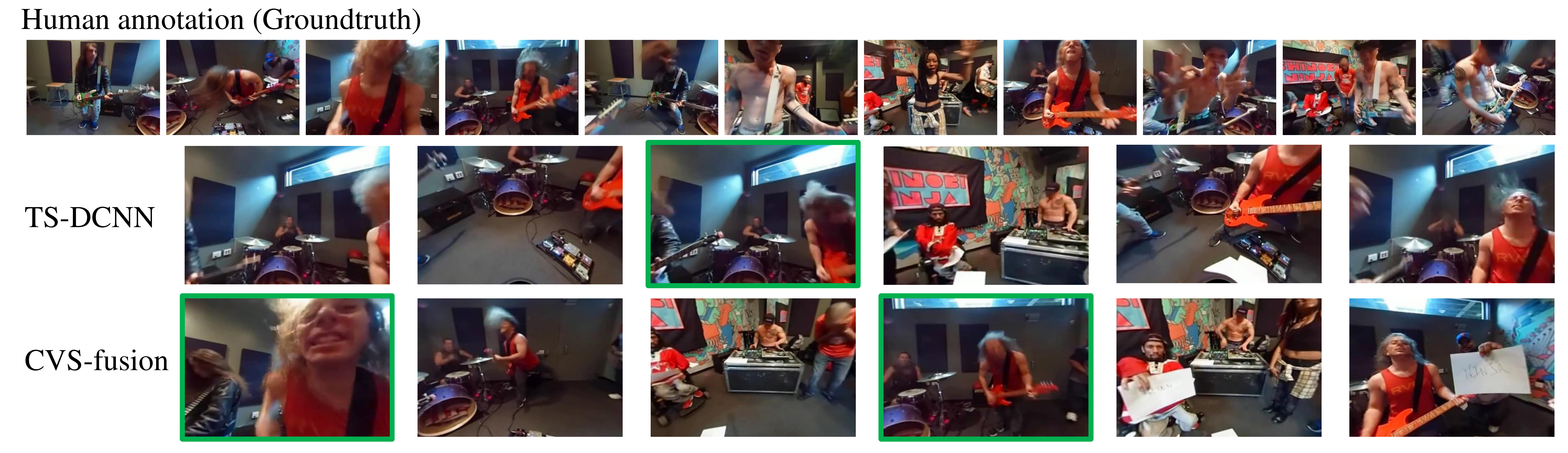}
	
	\caption{An example of highlight detection on a 360\degree~test video. We show human-annotated groundtruth highlight (top), and compare top-6 scored glimpses by TS-DCNN (middle) and our CVS model (bottom). Green boxes indicate true-positives. % (raise mAP).
	}
	\label{fig:projection_visualization}
\end{figure*}
%------------------------------------------------------------------------------

\subsection{Evaluation of the Pano2Vid Task}
\label{subsec:results_CVS}

For training of our CVS model, we use HumanCam positive samples $p_i$ and randomly cropped negative samples $n_i$ from panoramic videos in the Pano2Vid dataset. We use a simple max-margin loss $\mathcal{L}_i = \max(0, f(n_i) - f(p_i) + 1)$ instead of Eq.(\ref{eq:max-margin_loss}), because the training data are divided into two classes only (\ie positive and negative).
As an ablation study, we test our CVS model with three different configurations of video representation: (i) C3D only, (ii) Inception-v3 only, and (iii) both of C3D and Inception-v3. 
We also evaluate the variants of our method that use the smooth motion constraint, denoted by (*)-stitch. 
% To evaluate spatial summarization performance, we compare the predicted glimpses of each model with groundtruth.  

% \textbf{Quantitative Results}.
Table~\ref{tbl:results_mc_ret} shows the experimental results in terms of frame cosine similarity and overlap region metrics, which are official measures of the Pano2Vid task.
Our method CVS-Fusion-stitch outperforms all the baseline methods by a substantial margin in both metrics.
The smooth motion constraint helps better summarization as the variants denoted by (*)-stitch outperforms those without the constraint.
Even without the smooth motion constraint, the CVS models perform better than any baseline, regardless of which features the model uses.

\textbf{Computational cost}.
Table~\ref{tbl:computation_cost} compares the computation costs between our CVS and AutoCam. 
Due to the redundant computation of overlapping glimpses as shown in Figure \ref{fig:projection_vis}, AutoCam performs rectilinear projection on $\times 4.55$ of the actual area of the view sphere, 
while our framework projects $\times 1.96$ of the sphere area.    
Since the projection time is the bottleneck in practice, the computation time of our framework is significantly lower than that of AutoCam. 
We compare the average of processing time for one-minute 360\degree~video: 178 min of AutoCam and 11 min of CVS. 
We experiment on a machine with one Intel Xeon processor E5 2695 v4 (18 core) and GTX Titan X Pascal GPU.

\subsection{Evaluation of Highlight Detection}
\label{subsec:results_CVS}

\textbf{Quantitative results}.
Table~\ref{tbl:results_mAP} shows the results of highlight detection (\ie spatio-temporal summarization) on our 360\degree~video highlight dataset. 
We set $N=15$ as the highlight length for all algorithms. 
We do not test the AutoCam method, because it has no mechanism to generate a temporal summary; 
instead, we compare with pairwise ranking models, such as RankNet and TS-DCNN. 
Our CVS-Fusion achieves the best performance in terms of mAP.
As with observed performance drops, every key element of the CVS-Fusion model (\ie two different features and the triplet ranking loss) is critical to the performance.
Specifically, the triplet ranking loss significantly improves the performance of our model than the pairwise ranking loss, by providing a clearer guideline for a better composition.
For example, in terms of mAP for Wedding and MV datasets, our CVS model learned by the triplet ranking loss shows performance improvements by (5.98, 3.27), (4.73, 2.64), (3.62, 2.36), compared to RankNet, TS-DCNN, and CVS-Fusion-pairwise learned by the pairwise loss (\ie \textit{professional} and \textit{casual} as positives and \textit{random} as negatives), respectively.

\textbf{User studies via Amazon Mechanical Turk}.
We perform AMT tests to observe general users' preferences on the highlights detected by different algorithms. We randomly sample 20 test videos per topic from our 360\degree~video highlight dataset. At test, we show an original video and two sequences of highlight subshots generated by our model and one baseline method in a random order. Turkers are asked to pick a better one without knowing which comes from which method. We collect answers from three different turkers per test example. We compare our best method (CVS-Fusion) with three baselines: Center, RankNet, and TS-DCNN. % AutoCam-stitch, 

Table \ref{tbl:results_AMT} shows the results of AMT tests. It validates that general turkers prefer the output of our approach to those of baselines.
Note that our method using the triplet ranking loss is more preferred than the models using the pairwise ranking loss, RankNet and TS-DCNN.
These results coincide with quantitative results in Table~\ref{tbl:results_mAP}.

\textbf{Qualitative Results}.
Figure~\ref{fig:pos_overlap} shows qualitative examples of spatio-temporal glimpses that our CVS model chooses as highlights. % along with the corresponding composition score maps.
We also depict the composition score map, computed by Gaussian position-pooling, over the input video of the equirectangular projection (ERP) format.
We observe that high scores are often distributed to main characters, while relatively low scores are assigned to the regions with little saliency (\eg sky in the background).

Figure~\ref{fig:example_attention} illustrates the position score maps for (a) a professional, (b) a casual, and (c) a randomly sampled glimpses. 
In the example (a), our framework correctly assigns high scores to actual highlights of a music video characterized by the group dance with a good framing. 
In the example (b), our model attaches flat scores to the frames that capture proper content of the input video, but show a mediocre framing, shifted to the right. 
This may be due to low scores (depicted in blue) in the left vacant parts of the example $\{w^{0},w^{5},w^{10},w^{15},w^{20}\}$. 
This tendency is an incentive to move the view selection to the right to increase the score.
In the negative sample (c), all fitness scores are very low for the incorrect camera view.

Figure~\ref{fig:projection_visualization} shows an example of highlight detection on a 360\degree~test video of our dataset.
Our model successfully detects the main events labeled by human annotators, especially in top-6 scored glimpses.
Compared to TS-DCNN~\cite{yao:2016:CVPR}, the CVS model can successfully assign higher scores to the frames containing central events. 
By using both spatial and temporal features, our model can discover dynamic movements of main characters as highlights like a guitarist head-banging in MV, or important moments such as a couple kissing in Wedding.

\section{Conclusion}
\label{sec:conclusion}

We addressed a problem of 360\degree~video highlight detection via both spatial and temporal summarization.
% We also newly collected a 360\degree~video summarization dataset. 
We proposed a novel deep ranking model named  Composition View Score (CVS), which produces a spherical score map of composition per video segment to determine which view is suitable for highlight.
Using the spherical position score maps, our model is much faster at inference than existing methods. % (\eg more than $\times 16$ speed up than AutoCam). 
In our experiments, we showed that the CVS model outperformed state-of-the-art methods not only for spatial summarization in the Pano2Vid dataset, but also for highlight detection task in our newly collected video highlight dataset.

\smallskip 
\textbf{Acknowledgments}.
We thank Jinyoung Sung for the helpful discussion about the model. This work was supported by the Visual Display Business (RAK0117ZZ-21RF) of Samsung Electronics.
Gunhee Kim is the corresponding author.

\bibliographystyle{aaai}

\bibliography{aaai18_vrsumm} 
\end{document}